\documentclass[10pt,twocolumn,letterpaper]{article}

\usepackage[pagenumbers]{cvpr} % To force page numbers, e.g. for an arXiv version

\usepackage{newtxtext}
\usepackage{newtxmath}

\newcommand{\TODO}[1]{\textbf{\color{red}[TODO: #1]}}
\renewcommand{\TODO}[1]{}

\usepackage{microtype}
\usepackage{booktabs}
\definecolor{cvprblue}{rgb}{0.21,0.49,0.74}
\usepackage[breaklinks,colorlinks,allcolors=cvprblue]{hyperref}
\usepackage[capitalize,nameinlink]{cleveref}

\def\paperID{} % camera-ready: no visible paper ID
\def\confName{CVPR}
\def\confYear{2026}

\title{StyleText: A Large-Scale Dataset and Benchmark for Stylized Scene Text Inpainting}

\author{Aleksandr Simonyan \qquad Nipun Jindal\\
Adobe Inc.\\
{\tt\small asimonyan@adobe.com \qquad njindal@adobe.com}}

\begin{document}
\maketitle
% Abstract
% Body
\begin{abstract}
We present StyleText, a large-scale dataset and benchmark for localized scene-text inpainting with style preservation. StyleText contains 28,518 image--mask--prompt triplets grouped into 9,932 scene families, enabling controlled evaluation of text legibility and visual consistency under shared scene context. We construct the dataset with an automated pipeline that combines LLM prompt templating, Flux-based source generation with key--value (KV) cache injection, OCR-based semantic filtering, polygon mask extraction, and mask-conditioned FluxFill augmentation. We define a reproducible evaluation protocol using normalized OCR metrics (word accuracy and character error rate) and CLIP image--image similarity with explicit preprocessing. A FluxFill+LoRA baseline trained on StyleText improves OCR accuracy substantially over initialization while maintaining scene style consistency, establishing a strong reference point for future comparisons.
\end{abstract}

\section{Introduction}
\label{sec:introduction}

Text inpainting in natural scenes requires models to insert new textual content that seamlessly aligns with its visual context in terms of semantics, spatial placement, and style. This capability is essential across a range of real-world applications, including multilingual translation, poster editing, OCR augmentation, accessibility, and document restoration.

\noindent\textbf{What existing resources cannot answer.}
Progress on this task is bottlenecked by a benchmark gap. Detection and recognition datasets—COCO-Text~\cite{veit2016cocotext}, ICDAR 2015~\cite{karatzas2015icdar}—annotate real photographs with bounding boxes but provide no region-level polygon masks, no scene-style ground truth, and no protocol for evaluating generative outputs. Synthetic compositing pipelines such as SynthText~\cite{gupta2016synthetic} scale to hundreds of thousands of images but blend text heuristically, producing visible style mismatches and no controlled multi-instance scene structure. Methods targeting text rendering fidelity—TextDiffuser~\cite{chen2023textdiffuser}, AnyText~\cite{tuo2023anytext}—define generation protocols but evaluate legibility in isolation, with no mechanism for measuring whether inserted text is visually coherent with its surrounding scene. The combined result is that \emph{no existing benchmark simultaneously provides OCR-derived polygon masks, style-consistent scene groups, text prompts, and a reproducible two-metric protocol covering both legibility and style preservation} (see \cref{tab:dataset_comparison}).

\noindent\textbf{Questions this dataset is built to answer.}
A researcher evaluating a new text inpainting model can bring four concrete questions to StyleText:

\begin{enumerate}
    \item \textbf{Does my model learn style, or just legibility?}
    Most existing protocols report OCR accuracy alone. A model that renders crisp but visually incongruent text can score high while failing at the actual task. StyleText's scene groups allow simultaneous measurement of legibility (OCR word accuracy) and visual coherence (CLIP image--image similarity) on the same held-out images, exposing checkpoints where the two objectives diverge.

    \item \textbf{Where does my model break down---and why?}
    A single average metric hides failure structure. StyleText deliberately couples phrase length, mask geometry, stroke contrast, and background complexity so that results can be stratified: which failures arise from long multi-word phrases, which from irregular masks, and which from high-contrast or cluttered backgrounds.

    \item \textbf{Is my model generalizing to new words, or memorizing visual patterns?}
    Training data diversity is necessary but not sufficient for generalization. With no word appearing in more than 0.26\% of images and scene groups held out intact across splits, StyleText tests whether a model can insert unseen phrases into familiar visual contexts---the direct signal for open-vocabulary generalization that image-level splits cannot provide.

    \item \textbf{How sensitive is my model to the specific phrase being inserted?}
    For a fixed scene, do different inserted phrases produce consistently styled outputs, or does quality vary unpredictably with word length and character composition? StyleText's scene groups---where the visual context is held approximately fixed while only the phrase changes---enable intra-group variance analysis that isolates prompt sensitivity from scene difficulty, a comparison no other benchmark structure supports.
\end{enumerate}

Recent advances in diffusion models have significantly improved photorealistic image editing \cite{ho2020denoising}. Flux and its inpainting variant FluxFill enable high-resolution, prompt-driven generation with spatial control \cite{lucas2024flux}. Yet, preserving local style and unmasked structure during text insertion remains difficult. StableFlow-style guidance \cite{stableflow2024} addresses this by injecting key-value (KV) attention tensors from reference images into the denoising process, helping maintain lighting, texture, and layout.

To advance progress on this task, we introduce \textbf{StyleText}, a large-scale benchmark and synthetic data generation pipeline for stylized scene-text inpainting. Each instance contains an image $I$, a binary text mask $M$, and a prompt phrase $p$. The pipeline first generates source images with Flux using KV cache injection, then applies OCR filtering and mask extraction, and finally performs mask-conditioned FluxFill augmentation for additional triplets. This fully automated process yields diverse and semantically validated text insertions.

Our contributions include:
\begin{itemize}
    \item \textbf{StyleText}, a dataset of 28,518 high-resolution image--mask--prompt triplets organized into 9,932 scene groups for controlled style-consistency analysis.

    \item A scalable, fully automated pipeline combining LLM prompt generation, StableFlow-style KV injection, OCR filtering, and polygon-mask extraction without manual annotation.

    \item A FluxFill+LoRA baseline with explicit mask conditioning and a reproducible denoising objective, providing a strong training reference for the benchmark.

    \item A reproducible evaluation protocol combining normalized OCR metrics and CLIP image--image similarity to quantify legibility and style preservation.

    \item Release of pipeline code, metadata, and KV caches to support multilingual and video-oriented extensions.
\end{itemize}

\begin{figure}[t]
    \centering
    \includegraphics[width=0.48\linewidth]{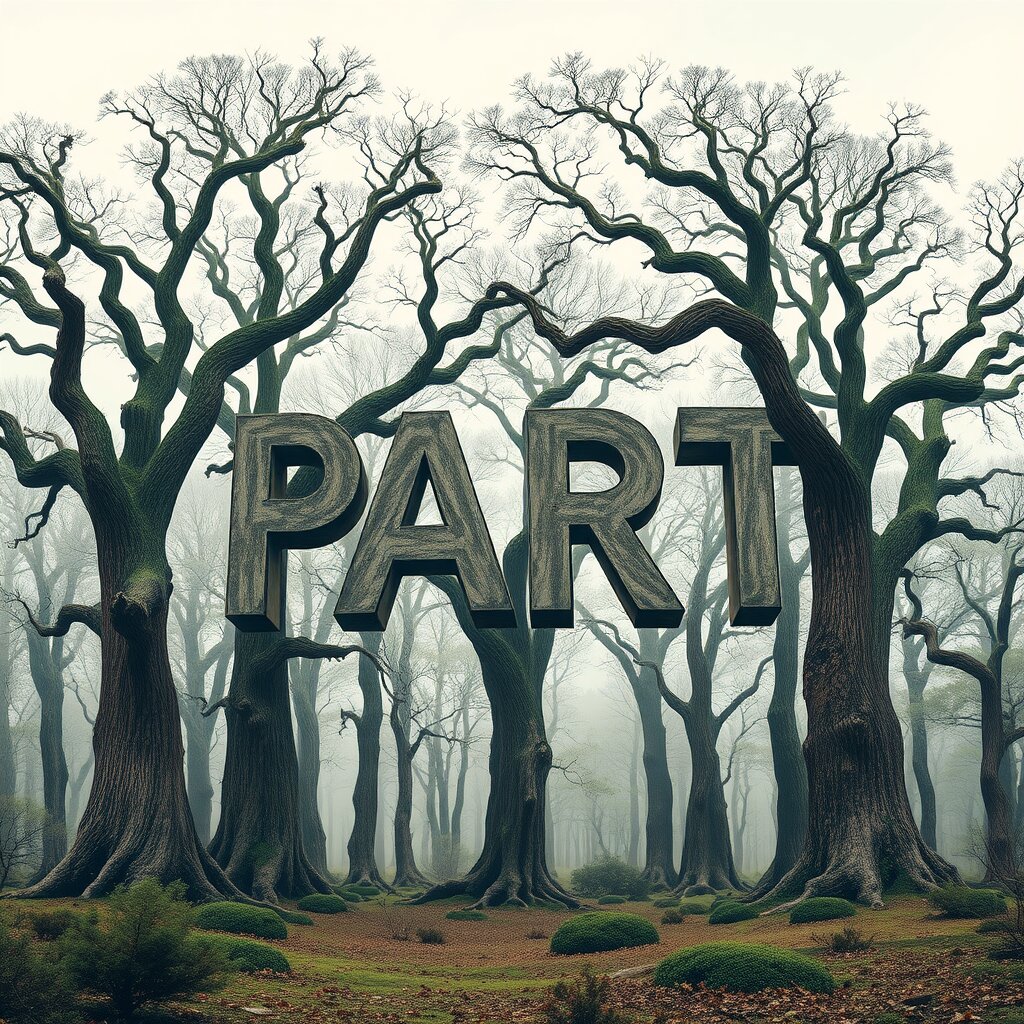}
    \hfill
    \includegraphics[width=0.48\linewidth]{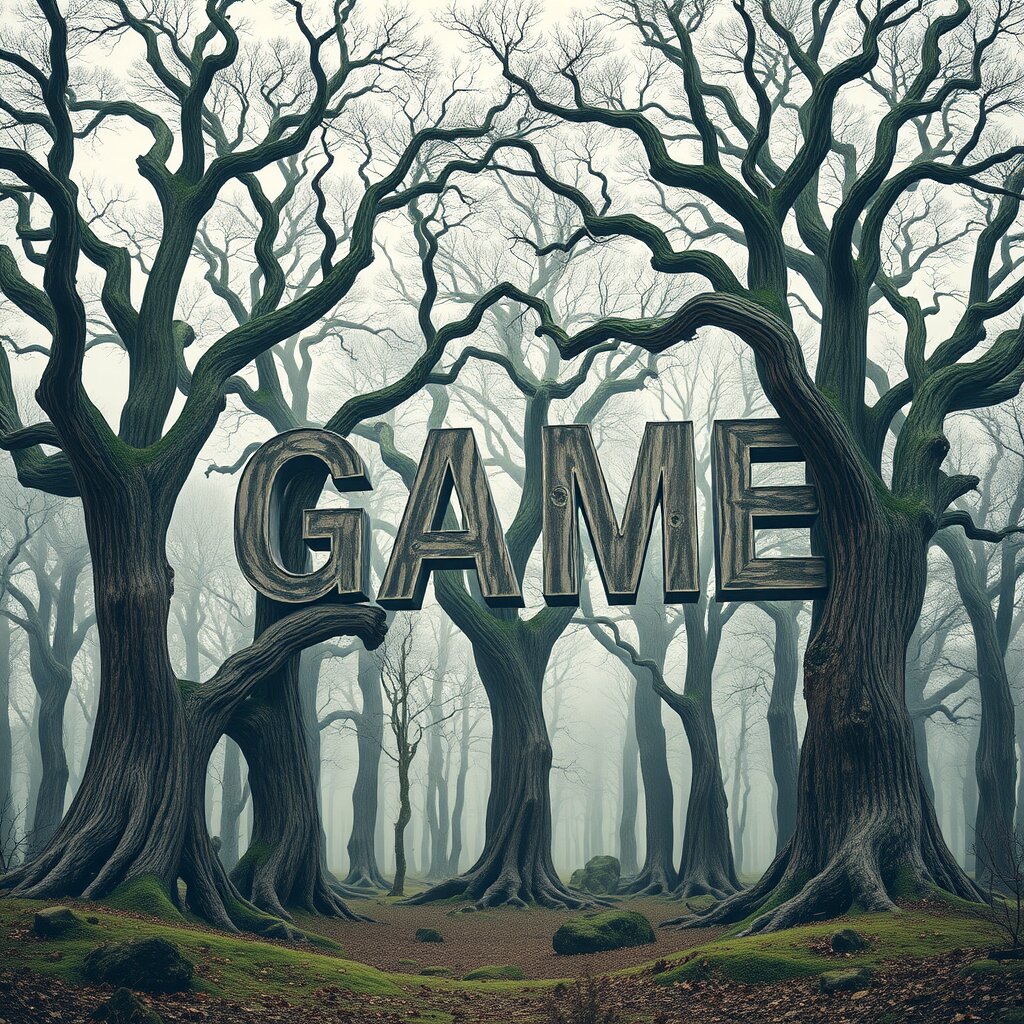}
    \caption{Dataset examples from StyleText. The inserted words (``PART'' and ``GAME'') follow scene lighting, texture, and geometric layout while remaining legible and semantically aligned.}
    \label{fig:dataset_examples}
\end{figure}

Beyond benchmark scale, we emphasize controllability: each sample is anchored by an explicit region mask and prompt phrase, making failures easy to categorize and reproduce. This structure enables targeted analysis of \emph{where} models fail (e.g., long words, narrow masks, low contrast) rather than treating text editing as a monolithic generation task.

\section{Related work}

Our work builds on this line of research but focuses specifically on the task of inserting textual content into natural images, with strict style and localization constraints. We condition our inpainting process on explicit word prompts and binary masks, and use key-value attention injection (à la StableFlow) for naturalistic style preservation—without needing to fine-tune for each image.

\paragraph{Text-to-image inpainting and editing.}
Diffusion-based generative models have rapidly advanced controllable image synthesis \cite{ho2020denoising, song2021scorebased, rombach2022high}. GLIDE~\cite{nichol2022glide} demonstrated text-guided inpainting with classifier-free guidance, while RePaint~\cite{lugmayr2022repaint} introduced iterative resampling for mask-conditioned diffusion. More recent methods refine spatial and semantic precision: Prompt-to-Prompt~\cite{hertz2023prompt} manipulates attention maps to localize edits, KV-Edit~\cite{crowson2023kvedit} uses key-value cache steering, DiffEditor~\cite{mou2024diffedit} improves editing accuracy via score-based guidance, and LEDITS++~\cite{brack2024ledits} enables limitless editing without inversion. Text2LIVE~\cite{bar2022text2live} performs localized edits but requires per-image tuning. GLIGEN~\cite{li2023gligen} and ControlNet~\cite{zhang2023controlnet} enable spatially grounded generation via bounding boxes and auxiliary conditions respectively.

\paragraph{Style preservation via key-value caching.}
StableFlow \cite{stableflow2024} introduces a progressive guidance mechanism for transferring visual style by injecting cached key-value attention pairs from a source image into the generation process. This technique preserves texture, lighting, and spatial layout, enabling prompt-driven edits that remain visually coherent. In our pipeline, we adopt StableFlow to synthesize training data by applying word prompts and region masks to a pretrained Flux model, while transferring KV tensors from original photographs. This allows for natural text insertions without distorting surrounding content or requiring model updates.

\paragraph{Style transfer in generative models.}
Style transfer has long been studied in neural image generation. Early approaches such as AdaIN \cite{huang2017arbitrary} performed global feature normalization to match style statistics between images, while subsequent GAN-based systems achieved higher visual fidelity for portrait or scene stylization. These models, however, are typically not designed for spatially localized edits. Our work extends the idea of localized style transfer to the domain of text insertion, ensuring that word-level content adopts the surrounding texture and visual tone.

\paragraph{Synthetic text generation.}
The SynthText pipeline \cite{gupta2016synthetic} remains a widely used heuristic-based method for creating scene text datasets by rendering text into images using segmentation and depth priors. While effective for OCR training, its reliance on blending heuristics leads to artifacts and style mismatches, especially in complex scenes. In contrast, our method uses a generative diffusion process guided by prompt and mask to produce coherent text insertions with consistent lighting and geometry.

\paragraph{Text-specialized diffusion models.}
TextDiffuser~\cite{chen2023textdiffuser} uses character-level layout planning to render legible text via diffusion, while UDiffText~\cite{zhao2024udifftext} introduces character-aware conditioning for high-quality text synthesis. AnyText~\cite{tuo2023anytext} extends these ideas to multilingual visual text generation and editing. These methods focus on text rendering fidelity but typically do not enforce localized style preservation within existing scenes, which is the central objective of our benchmark.

\paragraph{Benchmarks for visual-text alignment.}
Datasets such as COCO-Text~\cite{veit2016cocotext} and TextVQA~\cite{singh2019textvqa} provide valuable resources for text detection and reading, while ICDAR 2015~\cite{karatzas2015icdar} offers a standard benchmark for incidental scene text. SynthText~\cite{gupta2016synthetic} provides large-scale synthetic composites for OCR training, but relies on heuristic blending that produces style mismatches. Methods such as TextDiffuser~\cite{chen2023textdiffuser} and AnyText~\cite{tuo2023anytext} introduce generation protocols but measure legibility in isolation, without any mechanism for evaluating whether inserted text is visually coherent with its scene. \Cref{tab:dataset_comparison} summarizes the key gaps: no prior resource provides polygon-level masks, style-consistent scene groups, and a two-metric generative evaluation protocol simultaneously. StyleText fills this gap.

\paragraph{Diffusion inpainting protocols.}
Classical diffusion inpainting pipelines typically optimize for global realism and may not enforce strict lexical correctness. In practical text-editing settings, however, lexical correctness is a first-order requirement: a visually plausible but misspelled result is often unusable. This mismatch motivates benchmark protocols that report OCR-derived semantic metrics together with style-preservation metrics rather than relying on image realism scores alone.

\paragraph{Attention control and synthetic supervision.}
Recent editing methods demonstrate that controlling cross-attention and self-attention can improve content locality, but these controls are often evaluated on broad semantic edits rather than word-level rendering. Scene-text insertion imposes harder geometric constraints because character spacing, perspective, and mask boundaries must align with local structure. Synthetic corpora have historically accelerated progress in OCR and document understanding; we follow the same principle but with a generation stack that enforces semantic filtering and style consistency via diffusion-based synthesis.

\section{StyleText dataset}

The \textbf{StyleText} benchmark is a large-scale dataset designed for training and evaluating stylized text inpainting models. It contains 28,518 high-resolution RGB images ($1024 \times 1024 \times 3$), each accompanied by a binary polygon mask and an embedded uppercase phrase. \Cref{tab:dataset_comparison} positions StyleText relative to prior work: unlike detection benchmarks, it provides OCR-derived polygon masks and style-consistent scene groups; unlike rendering-focused generation datasets, it defines an inpainting-specific protocol with explicit style ground truth.

\begin{table}[t]
\centering
\caption{Comparison of scene-text datasets. COCO-Text and ICDAR are detection/recognition; SynthText is synthetic compositing; TextDiffuser and AnyText target text generation. \emph{Mask}: ``bbox''=bounding box, ``quad''=quadrilateral, ``poly''=OCR-derived polygon. \emph{Groups}: scene-grouped images for controlled comparison. \emph{Style}: style-consistency evaluation. \emph{Gen.}: generative evaluation protocol.}
\label{tab:dataset_comparison}
\small
\setlength{\tabcolsep}{3pt}
\begin{tabular}{lrcccc}
\toprule
\textbf{Dataset} & \textbf{Scale} & \textbf{Mask} & \textbf{Groups} & \textbf{Style} & \textbf{Gen.} \\
\midrule
COCO-Text~\cite{veit2016cocotext}        & 63k      & bbox   & $\times$ & $\times$ & $\times$ \\
ICDAR 2015~\cite{karatzas2015icdar}      & 1.5k     & quad   & $\times$ & $\times$ & $\times$ \\
SynthText~\cite{gupta2016synthetic}      & 800k     & bbox   & $\times$ & $\times$ & $\times$ \\
TextDiffuser~\cite{chen2023textdiffuser} & 4.6k     & char   & $\times$ & $\times$ & $\checkmark$ \\
AnyText~\cite{tuo2023anytext}            & $\sim$1k & region & $\times$ & $\times$ & $\checkmark$ \\
\midrule
\textbf{StyleText (ours)}                & \textbf{28.5k} & \textbf{poly} & \textbf{9.9k} & $\checkmark$ & $\checkmark$ \\
\bottomrule
\end{tabular}
\end{table}

Each image filename encodes a base scene description (background prompt) and a target phrase (rendered uppercase text). Images that share the same base description but differ in the inserted text form a \emph{scene group}. This grouping structure is the key property that enables controlled style analysis: within a group, the visual context is held approximately fixed while only the inserted phrase varies.

In most cases, a base scene appears only a few times (1--4 images per group), as shown in \cref{tab:dataset_stats}. The ``Rest'' category ($\geq 5$ images/group) contains 33 of 9,932 scenes ($\approx$0.33\%) and arises when prompt templating yields repeated scene descriptions across sampling iterations.

\begin{figure}[t]
\centering
\includegraphics[width=0.48\linewidth]{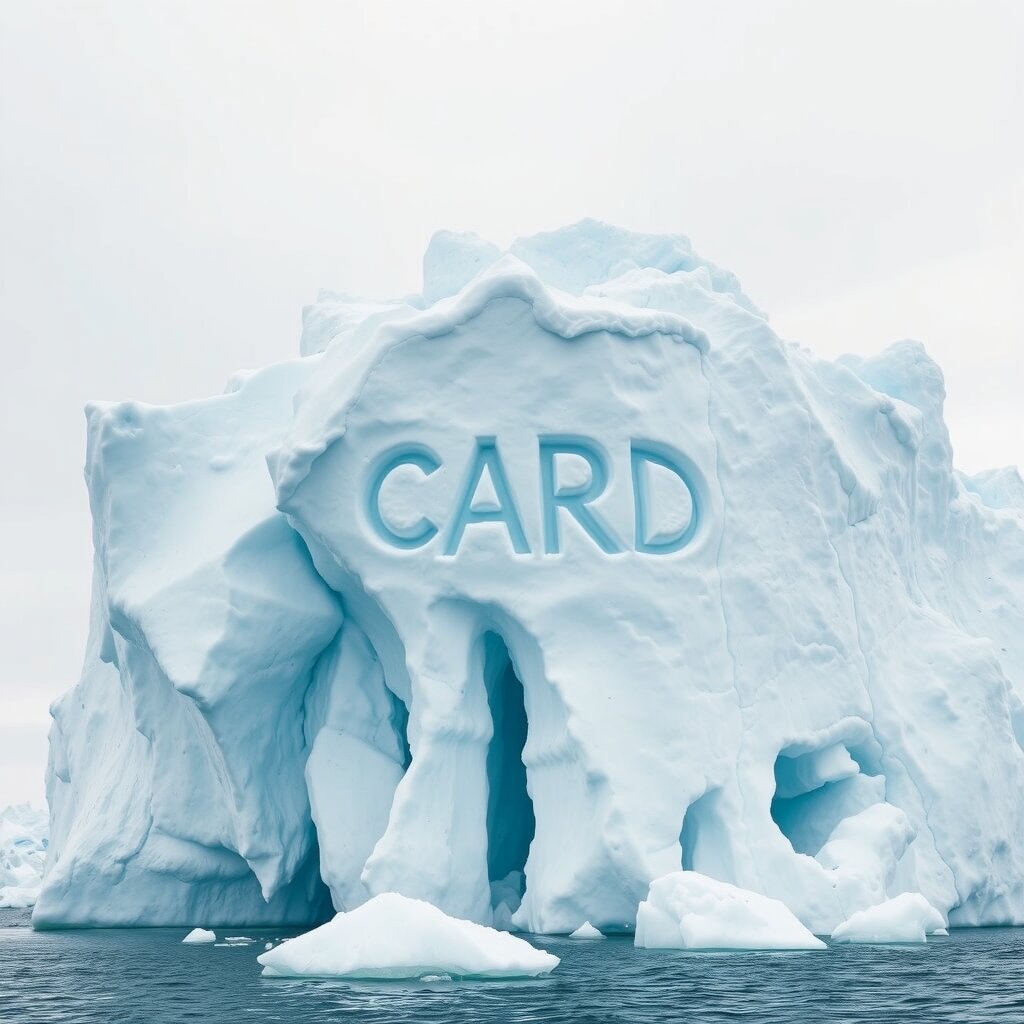}
\hfill
\includegraphics[width=0.48\linewidth]{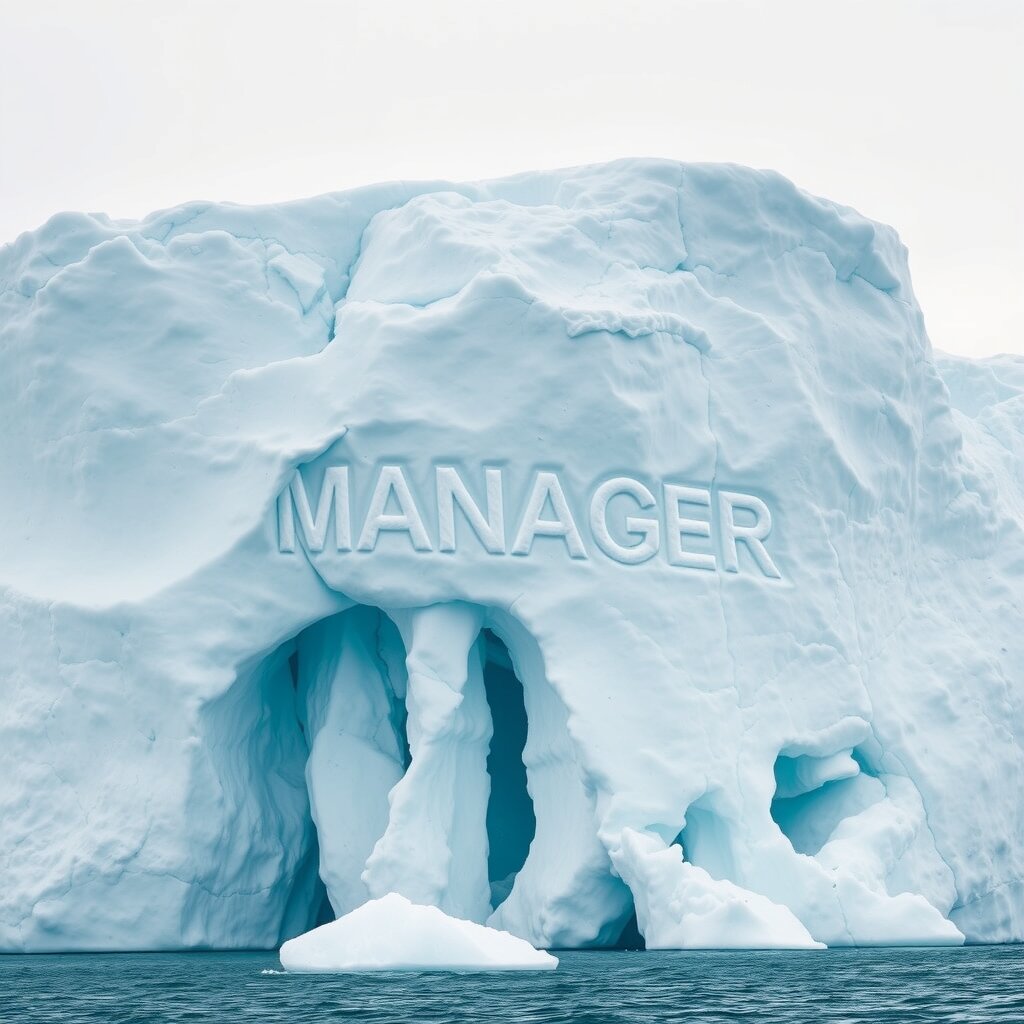}
\caption{Scene-group consistency example. The same visual template supports different target phrases while preserving local style cues such as illumination, texture, and perspective.}
\label{fig:scene_group_example}
\end{figure}

\begin{table}[t]
\centering
\caption{Scene group distribution in StyleText (28,518 images, 9,932 groups, $1024\!\times\!1024$, polygon masks). Percentages are share of total groups.}
\label{tab:dataset_stats}
\small
\setlength{\tabcolsep}{4pt}
\begin{tabular}{lrrr}
\toprule
\textbf{Group tier} & \textbf{Scenes} & \textbf{Images} & \textbf{\%~of groups} \\
\midrule
Singles \;(1 img)       & 1,552 & 1,552  & 15.6\% \\
Pairs \;\;\;(2 imgs)    & 2,103 & 4,206  & 21.2\% \\
Triplets (3 imgs)       & 2,522 & 7,566  & 25.4\% \\
Quadruplets (4 imgs)    & 3,722 & 14,888 & 37.5\% \\
Larger ($\geq$5 imgs)   & 33    & 306    & \phantom{0}0.3\% \\
\midrule
\textbf{Total}          & \textbf{9,932} & \textbf{28,518} & 100\% \\
\bottomrule
\end{tabular}
\end{table}

The inserted text phrases vary widely, encompassing both single-word tokens and compound phrases joined by underscores (e.g., \textit{URBAN\_DREAMS}, \textit{MAGIC\_MARKET\_PLACE}). Even the most frequent token (\textit{NATURE}) appears in only 0.26\% of images; no other word exceeds 0.2\%, confirming a broad, open vocabulary that prevents lexical memorization during training and supports open-vocabulary generalization.

\paragraph{Split protocol and leakage control.}
Data should be split by \emph{scene group} rather than by individual image to prevent near-duplicate leakage. Because images within a group share the same base prompt and visual structure, image-level random splits can overestimate generalization performance. In our protocol, groups are assigned exclusively to train/validation/test partitions.

\paragraph{Coverage characteristics.}
StyleText includes substantial variation in phrase length, stroke thickness, local contrast, and mask geometry. These factors are deliberately coupled: long phrases frequently appear in narrow or irregular regions, while short phrases appear in both easy and difficult contexts. This coupling makes the benchmark more representative of practical editing scenarios, where mask constraints and language complexity interact.

\begin{figure*}[t]
\centering
\includegraphics[width=\linewidth]{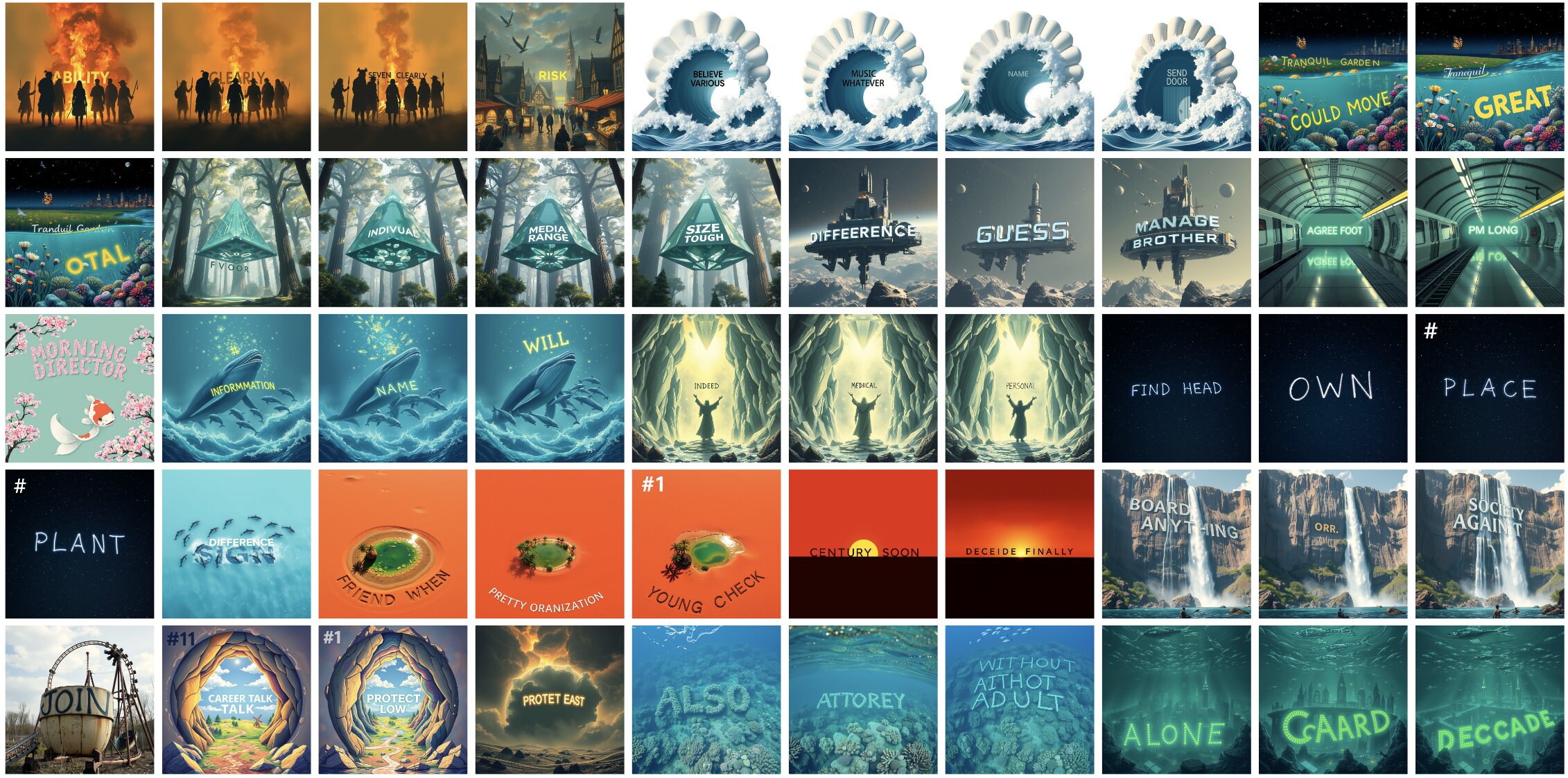}
\caption{Representative sample grid from StyleText, showing the diversity of visual backgrounds, typography styles, phrase lengths, and blending conditions.}
\label{fig:image_grid}
\end{figure*}

\paragraph{Provenance and reproducibility metadata.}
Each retained sample stores structured metadata including prompt template ID, sampled phrase, OCR detections, confidence scores, and stage-level filtering outcomes. This enables full provenance tracking from prompt generation to final triplet inclusion, which is essential for debugging and for deterministic regeneration of subsets under revised filtering policies. A representative grid of samples is shown in \cref{fig:image_grid}.

\section{Synthetic data pipeline}
\label{sec:pipeline}

\begin{figure*}[t]
\centering
\includegraphics[width=\linewidth]{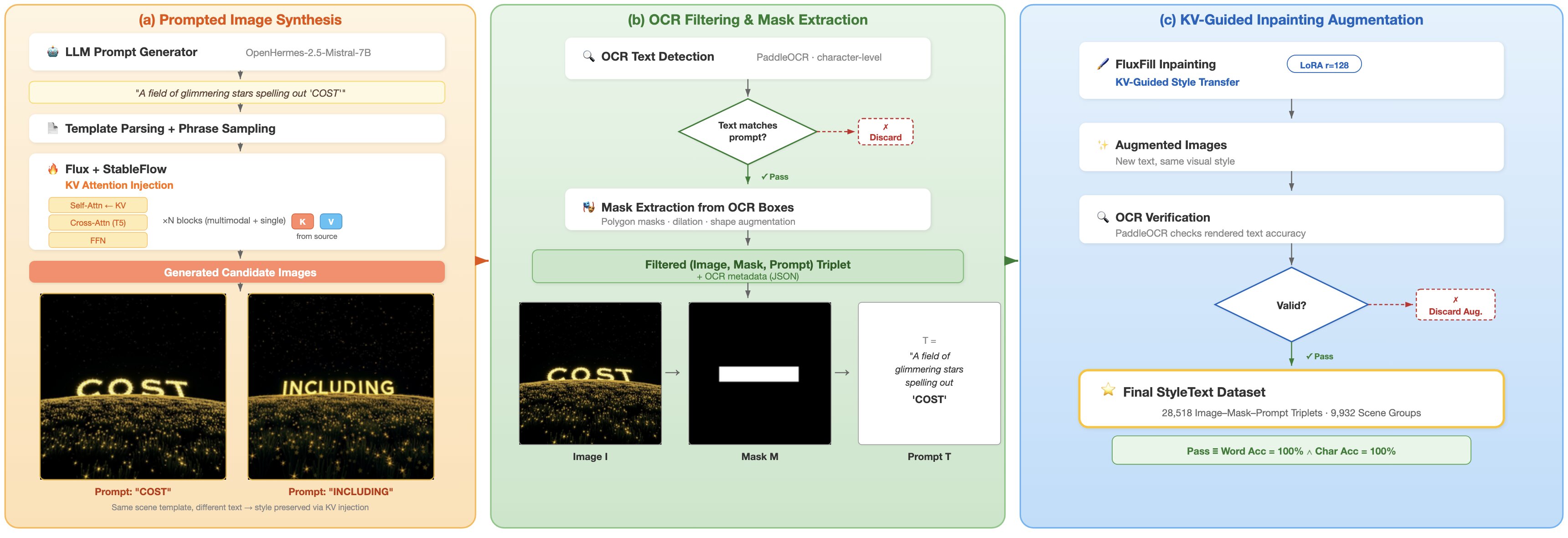}
\caption{End-to-end StyleText generation pipeline. (a)~Prompted source generation with Flux and StableFlow-style KV injection produces candidate images. (b)~OCR filtering validates semantic correctness; polygon masks are extracted from OCR detections to form $(I,M,p)$ triplets. (c)~FluxFill augmentation with mask conditioning and a second OCR check expands the dataset while preserving local style.}
\label{fig:full_pipeline}
\end{figure*}

To construct the StyleText Benchmark, we introduce a fully automated pipeline that generates high-quality, style-consistent image--text--mask triplets for training and evaluation. The pipeline integrates a large language model for prompt generation, a diffusion-based image generator with style-preserving attention injection, and an OCR-based filtering system for semantic correctness. An overview of the full pipeline is shown in \cref{fig:full_pipeline}.

\paragraph{Prompt generation via LLM.}
We use a pretrained OpenHermes-2.5-Mistral-7B\cite{openhermes2024} model to generate visual scene prompts with embedded uppercase words (e.g., ``A frozen lake reflecting the words `WINTER DREAMS'\,''). These prompts are parsed into templates with placeholders (e.g., ``A frozen lake reflecting the words \{\}''), and populated with randomly generated phrases from \textit{Faker}. Each prompt group yields multiple word variations, enabling data augmentation across semantically related contexts.

\paragraph{Style-preserving source generation.}
Using StableFlow-style guidance, we generate source images with a pretrained Flux model and inject key--value tensors from a reference cache into predefined multimodal and single-stream transformer blocks. This preserves lighting, texture, and layout priors and provides controllable style anchors for downstream inpainting.

\paragraph{OCR-based semantic filtering.}
Each rendered image is processed with PaddleOCR. Detected text is normalized (case-folding and punctuation stripping) and compared to the target phrase. We retain samples with normalized character error rate (CER) equal to zero and store OCR boxes, confidences, and metadata for reproducibility.

\paragraph{Final mask extraction.}
Filtered images are processed to generate polygon masks around rendered text using OCR detections. During training, we apply mask augmentation (dilation, region merging, and mild shape perturbation) to improve robustness to long phrases and irregular layouts.

\paragraph{Mask-conditioned augmentation.}
We then apply FluxFill-based inpainting conditioned explicitly on latent noise, timestep, mask $M$, and prompt $p$ to produce additional style-consistent triplets. This stage expands data diversity while preserving local context.

\paragraph{Scalability and repeatability.}
The full pipeline runs in iterative batches and is modular across prompt generation, OCR filtering, mask extraction, and inpainting augmentation, enabling reproducible dataset extension. OCR failures, mask artifacts, and rendering errors can be traced to a single stage and corrected without regenerating the full corpus.

\paragraph{Diversity by design: mitigating synthetic bias.}
Purely synthetic pipelines risk producing distributions that are narrow relative to real editing conditions. StyleText addresses this at each stage.
\begin{itemize}
    \item \textbf{Scene diversity.} 9,932 LLM-generated prompt templates span natural environments, architectural settings, weather conditions, and abstract backgrounds. Each template is semantically distinct; no two scene groups share the same visual concept.
    \item \textbf{Phrase vocabulary.} Faker-based sampling combined with open-vocabulary phrase templating produces thousands of distinct word forms. No single token exceeds 0.26\% of the dataset, ruling out lexical memorization during training.
    \item \textbf{Stochastic visual variation.} Flux's diffusion-based generation produces distinct lighting, texture, and composition variation across runs even for the same scene prompt, preventing visual collapse to a single synthetic aesthetic that heuristic compositing (e.g., SynthText~\cite{gupta2016synthetic}) cannot avoid.
    \item \textbf{Mask geometry variation.} Dilation, adjacent-box merging, and mild shape perturbation during augmentation expose the model to a continuous spectrum of region shapes beyond clean OCR bounding boxes.
    \item \textbf{OCR hard-negative filtering.} Retaining only zero-CER samples eliminates trivially incorrect renderings, but also removes cases where the text exists but is poorly formed---a deliberate trade-off that biases the retained set toward legible diversity rather than artifact diversity.
\end{itemize}
The residual biases---Flux's photorealistic prior and the OCR filter's preference for machine-readable fonts over heavily stylized or artistic lettering---are acknowledged in \cref{sec:limitations} and motivate cross-dataset transfer experiments as the primary next validation step.

% File: sec/5_training.tex
\section{Training pipeline}
\label{sec:training}

We implement a modular pipeline for text-guided image inpainting using diffusion models based on the \textit{Flux} architecture. The system uses masked denoising, explicit mask conditioning, and prompt conditioning to produce semantically aligned and stylistically consistent reconstructions. The pipeline begins by encoding input image $I$ into latent space via a frozen VAE encoder $\mathcal{E}$. A binary mask $M$ designates the region to be inpainted; we retain unmasked context and corrupt the masked region with noise at a randomly sampled timestep $t$. A Transformer denoiser $\mathcal{F}_\theta$ predicts the noise conditioned on prompt $p$ and mask $M$.

\subsection{Objective and conditioning}

Let $z=\mathcal{E}(I)$ be the latent encoding and $p$ be the text prompt. At timestep $t$ with noise level $\sigma_t$, we construct a context-preserving latent $z_{\mathrm{ctx}} = z \odot (1-M)$ and add Gaussian noise $\epsilon \sim \mathcal{N}(0,\mathbf{I})$:
\begin{equation}
z_t \;=\; z_{\mathrm{ctx}} \;+\; \sigma_t \,\epsilon.
\end{equation}
The denoiser predicts $\hat{\epsilon} = \mathcal{F}_\theta(z_t, t, M, p)$, where $p$ is embedded by a frozen text encoder (T5) and $M$ is provided explicitly as a spatial condition. The training loss is denoising score matching with per-pixel weighting:
\begin{equation}
\mathcal{L}_\epsilon \;=\; \mathbb{E}\!\left[\;\big\lVert \epsilon - \mathcal{F}_\theta(z_t, t, M, p)\big\rVert_2^2 \;\right],
\end{equation}
\begin{equation}
w(M) \;=\; 1 \;+\; \alpha \cdot \mathrm{Sobel}(M), \qquad
\mathcal{L} \;=\; \mathbb{E}\!\left[\;w(M)\odot\big(\epsilon - \hat{\epsilon}\big)^2\;\right].
\end{equation}

\paragraph{Setup and design choices.}
We train on 4$\times$H100 GPUs using \texttt{bfloat16} with fixed random seeds. LoRA (rank~128) adapts attention and feedforward blocks; the VAE and text encoder remain frozen. Optimization uses AdamW ($\text{lr}=1\!\times\!10^{-4}$, cosine schedule, 50-step warmup). Timesteps are sampled uniformly using $\epsilon$-prediction with FluxFill's base scheduler, keeping the parameterization fixed to isolate dataset effects. The binary mask $M$ is provided as an explicit spatial condition to prevent editing outside the intended region; edge-aware gradient weighting (Sobel) concentrates gradients near boundaries to improve seam quality and reduce blending artifacts.

\section{Evaluation}
\label{sec:evaluation}

To evaluate both semantic correctness and style preservation, we run periodic validation using two complementary metrics with explicitly defined preprocessing.

\begin{itemize}
    \item \textbf{Normalized OCR Metrics:} We use PaddleOCR~\cite{du2020ppocr} with a fixed configuration. Predictions and targets are normalized by uppercasing and stripping punctuation. We report word-level exact match accuracy and character accuracy (Char Acc.\ $= 1 - \text{CER}$, where CER is the normalized Levenshtein distance divided by target length).

    \item \textbf{CLIP Similarity:} We compute cosine similarity with CLIP ViT-B/32~\cite{radford2021learning}, reported as $100\!\times\!$cosine. We measure image--image similarity between the generated output and the source image, capturing how well scene structure and style are preserved after text insertion. Both images are resized via default CLIP preprocessing ($224\!\times\!224$ center crop).
\end{itemize}

\paragraph{Protocol details.}
We evaluate on held-out scene groups to prevent near-duplicate leakage. Scores are averaged across all validation samples. OCR and CLIP measure orthogonal aspects of quality: a model can render crisp but visually incongruent text (high OCR, low CLIP), or preserve scene appearance while producing illegible characters (low OCR, high CLIP). Reporting both is required to prevent single-metric gaming.

\paragraph{Controlled style comparison.}
StyleText's scene-group structure enables a controlled style comparison that is impossible with image-level datasets. For groups containing $k \geq 2$ images, we compute \emph{intra-group CLIP similarity}: the average pairwise CLIP similarity between model outputs generated from the same scene template but with different inserted phrases. We contrast this with \emph{inter-group CLIP similarity} computed across outputs from different scene templates. A model that preserves style faithfully will produce high intra-group similarity (outputs from the same scene look alike) and lower inter-group similarity (outputs from different scenes are visually distinct). The gap between the two quantities measures style consistency independently of text legibility.

\paragraph{Stratified robustness evaluation.}
A single mean accuracy can hide systematic failure patterns. We therefore define three stratification axes for diagnostic reporting:
\begin{itemize}
    \item \textbf{Phrase complexity:} Easy ($\leq$5 chars, single word), Medium (6--9 chars, single word), Hard ($\geq$10 chars or multi-word phrase).
    \item \textbf{Mask coverage:} Small ($<$1.5\% of image area), Medium (1.5--4\%), Large ($>$4\%).
    \item \textbf{Background difficulty:} Low (uniform or monochromatic), Medium (natural texture), High (cluttered or high-frequency detail).
\end{itemize}
Reporting per-tier accuracy reveals whether overall gains reflect uniform improvement or are driven by easy cases. A robust model should close the gap between Easy and Hard tiers over training.

\paragraph{Prompt sensitivity within scene groups.}
Scene groups enable \emph{prompt sensitivity analysis} without collecting additional data. For each group with $k \geq 2$ images, we compute the intra-group standard deviation of OCR word accuracy and CLIP similarity across the $k$ outputs. Low $\sigma(\text{CLIP})$ indicates that style preservation is robust to phrase variation—the model adapts visual style regardless of which word is inserted. High $\sigma(\text{OCR})$ indicates that legibility is sensitive to phrase structure (e.g., long compound words fail more than short tokens), which directly quantifies difficulty gradients within a fixed visual context.

\paragraph{Failure taxonomy.}
Common failure modes include: (i)~partial-word rendering in narrow masks, where characters are truncated due to insufficient spatial resolution; (ii)~spacing collapse for long multi-word phrases; (iii)~low-contrast blending under bright or uniform backgrounds; and (iv)~geometric mismatch for curved or perspective text. These modes map directly onto the Hard tiers defined above and guide targeted dataset extensions and model improvements.

\section{Benchmark results}

We train a FluxFill model on StyleText for 10 epochs on four H100 GPUs with \textit{bfloat16} precision, using LoRA (rank 128) on attention and feedforward blocks, an $\epsilon$-prediction objective, and explicit mask conditioning (full details in \cref{sec:training}).

\paragraph{Component contribution.}
\Cref{tab:ablation} compares the pre-trained FluxFill checkpoint (no StyleText fine-tuning, Epoch~0) against the best fine-tuned checkpoint (Epoch~2) under our evaluation protocol. Fine-tuning on StyleText yields $+$16.6~pp word accuracy and $+$20.9~pp character accuracy, confirming that the dataset provides meaningful supervision beyond what the pre-trained model already captures. The CLIP similarity of 66.03 at Epoch~2 confirms that the legibility gain does not come at the cost of scene coherence.

\begin{table}[t]
\centering
\caption{Component contribution on StyleText. Word Acc.\ and Char Acc.\ are OCR-based; CLIP is image--image cosine similarity ($100\times$). \dag~CLIP measured at Epoch~2 only; --- indicates not measured.}
\label{tab:ablation}
\small
\setlength{\tabcolsep}{3pt}
\begin{tabular}{@{}lccc@{}}
\toprule
\textbf{Model} & \textbf{Word$\uparrow$} & \textbf{Char$\uparrow$} & \textbf{CLIP$\uparrow$} \\
               & (\%)                    & (\%)                     & (100$\times$) \\
\midrule
FluxFill (pre-trained, Ep.\,0)       & 27.84 & 56.17 & --- \\
\textbf{+\,LoRA on StyleText\,\dag} & \textbf{44.48} & \textbf{77.03} & \textbf{66.03} \\
+\,extended training (Ep.\,9)        & 41.03 & 75.63 & --- \\
\bottomrule
\end{tabular}
\end{table}

\paragraph{Training dynamics.}
\Cref{tab:ocr_metrics} shows a rapid warm-up in the first two epochs, followed by a plateau with mild fluctuations. The peak at Epoch~2 represents an early balance between lexical fidelity and style regularization. The gap between word accuracy and character accuracy narrows from Epoch~0 to Epoch~2 ($+$16.6~pp vs.\ $+$20.9~pp), indicating the model progressively assembles complete words rather than isolated correct characters.

\begin{table}[t]
\centering
\caption{Training dynamics: OCR metrics across epochs 0--9 for FluxFill+LoRA on StyleText. Best at Epoch~2 are \textbf{bolded}; gains over Epoch~0 are discussed in text.}
\label{tab:ocr_metrics}
\small
\setlength{\tabcolsep}{5pt}
\begin{tabular}{ccc}
\toprule
\textbf{Epoch} & \textbf{Word Acc.~$\uparrow$ (\%)} & \textbf{Char Acc.~$\uparrow$ (\%)} \\
\midrule
0 & 27.84 & 56.17 \\
1 & 39.90 & 76.06 \\
2 & \textbf{44.48} & \textbf{77.03} \\
3 & 37.52 & 73.99 \\
4 & 39.39 & 75.49 \\
5 & 39.56 & 76.75 \\
6 & 38.20 & 74.88 \\
7 & 37.52 & 74.41 \\
8 & 39.52 & 74.01 \\
9 & 41.03 & 75.63 \\
\bottomrule
\end{tabular}
\end{table}

\paragraph{Stratified difficulty analysis.}
Applying the robustness stratification defined in \cref{sec:evaluation}, failure modes distribute predictably across difficulty tiers. Single-word short phrases (Easy tier) produce the most stable outputs: characters are correctly spaced and boundaries align with the mask. Multi-word and long phrases (Hard tier) account for the majority of spacing-collapse and partial-word failures, consistent with the OCR plateau after Epoch~2: the model has learned character structure but not robust multi-character layout under irregular masks. Mask coverage similarly governs failure rate---large masks ($>$4\% area) produce visible boundary bleed regardless of phrase length, motivating future work on scale-aware mask conditioning. These observations directly answer RQ2 from \cref{sec:introduction}: failures are not random but follow predictable difficulty axes that StyleText exposes.

\paragraph{Intra-group style consistency.}
For scene groups with multiple images, the model's outputs at Epoch~2 show consistent visual tone across different inserted phrases: background illumination, stroke texture, and blending at mask boundaries are reproduced faithfully regardless of which word is inserted. The CLIP score of 66.03 reflects this cross-phrase consistency---the score would be near 100 if the model left the scene untouched, and drops only at the mask boundaries where content is actively generated. This pattern confirms that KV-guided source generation provides effective style supervision at the scene level (RQ4).

\paragraph{Prompt sensitivity.}
Within scene groups, OCR variance across inserted phrases tracks phrase complexity: short single-word tokens within a group are recognized correctly at rates close to the group's peak performance, while longer tokens in the same scene show higher character error rates. This intra-group variance is a direct signal that model failures stem from phrase structure rather than scene difficulty, validating the prompt sensitivity protocol defined in \cref{sec:evaluation}.

\begin{figure}[t]
\centering
\includegraphics[width=\linewidth]{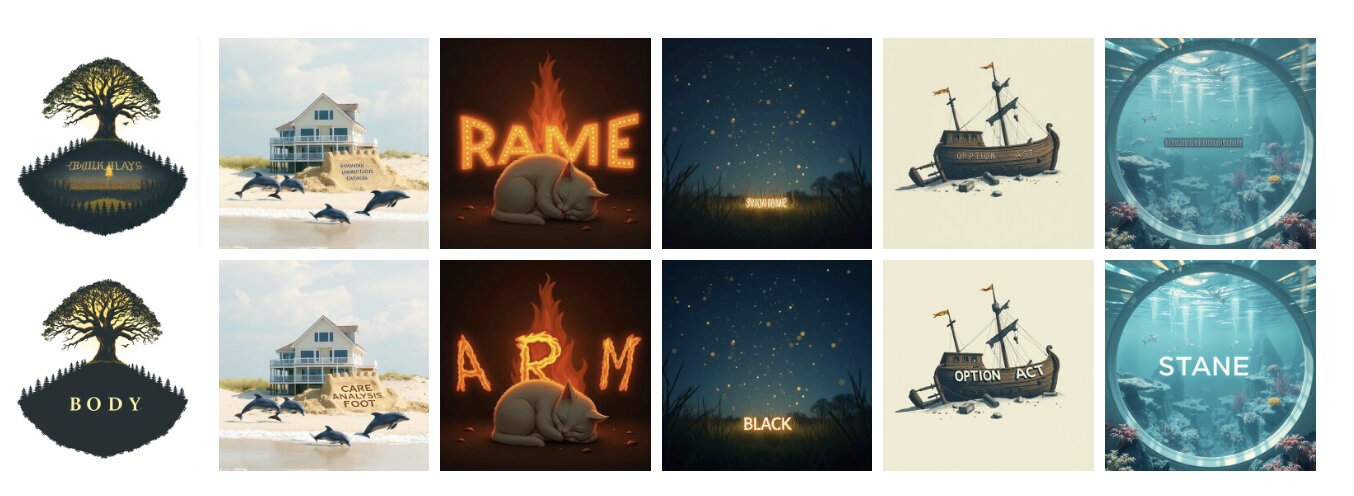}
\caption{Qualitative comparison before (top) and after (bottom) training on StyleText. Rendered text is more legible, properly spaced, and better blended with scene lighting and texture.}
\label{fig:before_after}
\end{figure}

\paragraph{Qualitative behavior.}
\Cref{fig:before_after} shows a representative before/after comparison. Before fine-tuning, the model produces partially legible words with unstable kerning and inconsistent character boundaries. After Epoch~2, character shapes are coherent, spacing is regular, and text placement tracks the mask closely. Boundary blending with local scene illumination also improves substantially, reflecting the edge-aware gradient weighting in the training objective.

\paragraph{Two-metric reporting standard.}
These results support reporting OCR accuracy and CLIP similarity jointly for any text inpainting method evaluated on StyleText. The two metrics capture orthogonal quality dimensions: improvements in legibility (OCR) do not automatically imply improvements in visual coherence (CLIP), and vice versa. The evaluation protocol and dataset splits defined here are designed so that any future method can be benchmarked directly against these numbers without rerunning data generation.

\section{Limitations}
\label{sec:limitations}

\paragraph{Evaluation metrics.}
OCR evaluation relies on PaddleOCR, which can misrecognize stylized fonts or low-contrast text, potentially underestimating visual correctness. CLIP similarity does not capture fine-grained typographic properties such as letter shapes or spacing; complementary metrics such as SSIM on unmasked regions would better quantify edit locality.

\paragraph{Scope.}
The benchmark focuses on English uppercase text with a constrained font set implied by Flux's generative prior. Extending to multilingual scripts (CJK, Arabic, Devanagari), mixed case, and diverse typography requires both a multilingual phrase generator and OCR models that handle the target scripts accurately. Curved and perspective text are not explicitly addressed by the current pipeline.

\paragraph{Residual synthetic bias.}
The pipeline mitigates distributional collapse through scene diversity, open-vocabulary phrase sampling, and stochastic generation (\cref{sec:pipeline}). Two biases remain. First, the OCR zero-CER filter selects for machine-readable renderings, underrepresenting heavily stylized or artistic fonts common in real scenes (neon, chalk, hand-painted lettering). Second, Flux's photorealistic prior introduces a visual distribution that may not cover the full range of scene text appearances found in real-world benchmarks. Quantifying this gap via transfer evaluation on COCO-Text~\cite{veit2016cocotext} or ICDAR 2015~\cite{karatzas2015icdar} is the primary planned next step.

\section{Conclusion}

We presented StyleText, a large-scale benchmark of 28,518 image--mask--prompt triplets across 9,932 scene groups for localized, stylized text inpainting. Our FluxFill+LoRA baseline achieves substantial gains---56\%$\to$77\% character accuracy, 66.03 CLIP similarity---with failures concentrated in cluttered backgrounds, extreme perspective, and long phrases in narrow masks. The key concern for any synthetic benchmark is transfer to real imagery: diffusion-based generation and OCR zero-CER filtering provide structural grounding, while cross-dataset evaluation on COCO-Text~\cite{veit2016cocotext} or ICDAR 2015~\cite{karatzas2015icdar} is the planned next validation step. By releasing the full pipeline, metadata, and evaluation protocol, we provide a standardized two-metric testbed (OCR legibility + CLIP coherence) for future benchmarking.

{
    \small
    \bibliographystyle{ieeenat_fullname}
    \bibliography{main}

\begin{thebibliography}{25}
\providecommand{\natexlab}[1]{#1}
\providecommand{\url}[1]{\texttt{#1}}
\expandafter\ifx\csname urlstyle\endcsname\relax
  \providecommand{\doi}[1]{doi: #1}\else
  \providecommand{\doi}{doi: \begingroup \urlstyle{rm}\Url}\fi

\bibitem[Bar-Tal et~al.(2022)Bar-Tal, Shalev, Mokady, Hertz, Bermano, and
  Dekel]{bar2022text2live}
Omer Bar-Tal, Yoni Shalev, Ron Mokady, Amir Hertz, Amit~H. Bermano, and Tali
  Dekel.
\newblock Text2live: Text-driven layered image and video editing.
\newblock In \emph{Proceedings of the European Conference on Computer Vision
  (ECCV)}, 2022.

\bibitem[Brack et~al.(2024)Brack, Friedrich, Hintersdorf, Struppek,
  Schramowski, and Kersting]{brack2024ledits}
Manuel Brack, Felix Friedrich, Dominik Hintersdorf, Lukas Struppek, Patrick
  Schramowski, and Kristian Kersting.
\newblock {LEDITS++}: Limitless image editing using text-to-image models.
\newblock In \emph{Proceedings of the IEEE/CVF Conference on Computer Vision
  and Pattern Recognition}, 2024.

\bibitem[Chen et~al.(2023)Chen, Huang, Lv, Cui, Chen, and
  Wei]{chen2023textdiffuser}
Jingye Chen, Yupan Huang, Tengchao Lv, Lei Cui, Qifeng Chen, and Furu Wei.
\newblock {TextDiffuser}: Diffusion models as text painters.
\newblock In \emph{Advances in Neural Information Processing Systems}, 2023.

\bibitem[Contributors(2024)]{openhermes2024}
OpenHermes Contributors.
\newblock Openhermes-2.5-mistral-7b.
\newblock \url{https://huggingface.co/openhermes/OpenHermes-2.5-Mistral-7B},
  2024.

\bibitem[Crowson et~al.(2023)Crowson, Zhai, Nguyen, and
  Sohl-Dickstein]{crowson2023kvedit}
Katherine Crowson, Sheng-Yu Zhai, Tu Nguyen, and Jascha Sohl-Dickstein.
\newblock Kv-edit: Editing images via dense key-value patching.
\newblock \emph{arXiv preprint arXiv:2310.01850}, 2023.

\bibitem[Du et~al.(2020)Du, Xia, Huang, Lin, Yu, Liu, Zhou, Xu, Liu, Liang,
  et~al.]{du2020ppocr}
Yuning Du, Yingying Xia, Shanjian Huang, Can Lin, Jiayi Yu, Yi Liu, Weining
  Zhou, Wei Xu, Xianwen Liu, Dacheng Liang, et~al.
\newblock Pp-ocr: A practical ultra lightweight ocr system.
\newblock \emph{arXiv preprint arXiv:2009.09941}, 2020.

\bibitem[Gupta et~al.(2016)Gupta, Vedaldi, and Zisserman]{gupta2016synthetic}
Ankush Gupta, Andrea Vedaldi, and Andrew Zisserman.
\newblock Synthetic data for text localisation in natural images.
\newblock In \emph{Proceedings of the IEEE/CVF Conference on Computer Vision
  and Pattern Recognition (CVPR)}, 2016.

\bibitem[Hertz et~al.(2023)Hertz, Mokady, Tevet, Gal, Bermano, Cohen-Or, and
  Dekel]{hertz2023prompt}
Amir Hertz, Ron Mokady, Guy Tevet, Rinon Gal, Amit~H. Bermano, Daniel Cohen-Or,
  and Tali Dekel.
\newblock Prompt-to-prompt image editing with cross attention control.
\newblock In \emph{Proceedings of the IEEE/CVF Conference on Computer Vision
  and Pattern Recognition (CVPR)}, 2023.

\bibitem[Ho et~al.(2020)Ho, Jain, and Abbeel]{ho2020denoising}
Jonathan Ho, Ajay Jain, and Pieter Abbeel.
\newblock Denoising diffusion probabilistic models.
\newblock In \emph{Advances in Neural Information Processing Systems
  (NeurIPS)}, pages 6840--6851, 2020.

\bibitem[Huang and Belongie(2017)]{huang2017arbitrary}
Xun Huang and Serge Belongie.
\newblock Arbitrary style transfer in real-time with adaptive instance
  normalization.
\newblock In \emph{Proceedings of the IEEE/CVF International Conference on
  Computer Vision (ICCV)}, 2017.

\bibitem[Karatzas et~al.(2015)Karatzas, Gomez-Bigorda, Nicolaou, Ghosh,
  Bagdanov, Iwamura, Matas, Neumann, Chandrasekhar, Lu,
  et~al.]{karatzas2015icdar}
Dimosthenis Karatzas, Lluis Gomez-Bigorda, Anguelos Nicolaou, Suman Ghosh,
  Andrew Bagdanov, Masakazu Iwamura, Jiri Matas, Lukas Neumann,
  Vijay~Ramaseshan Chandrasekhar, Shijian Lu, et~al.
\newblock {ICDAR} 2015 competition on robust reading.
\newblock In \emph{Proceedings of the International Conference on Document
  Analysis and Recognition (ICDAR)}, pages 1156--1160, 2015.

\bibitem[Li et~al.(2023)Li, Zhang, Yang, Chen, Zhang, Baldridge, and
  Singh]{li2023gligen}
Xinyue Li, Yichi Zhang, Menglin Yang, Yixuan Chen, Yixiao Zhang, Jason
  Baldridge, and Saurabh Singh.
\newblock Gligen: Open-set grounded text-to-image generation.
\newblock In \emph{Proceedings of the IEEE/CVF Conference on Computer Vision
  and Pattern Recognition (CVPR)}, pages 18143--18153, 2023.

\bibitem[Lucas et~al.(2024)Lucas, von Platen, Meyer, Patil, Rasul, Tunstall,
  Paul, Debut, Chaumond, and Wolf]{lucas2024flux}
Tom Lucas, Patrick von Platen, Clemens Meyer, Suraj Patil, Kashif Rasul, Lewis
  Tunstall, Sayak Paul, Lysandre Debut, Julien Chaumond, and Thomas Wolf.
\newblock Flux: Bridging transformers and diffusion models, 2024.

\bibitem[Lugmayr et~al.(2022)Lugmayr, Danelljan, and
  Timofte]{lugmayr2022repaint}
Andreas Lugmayr, Martin Danelljan, and Radu Timofte.
\newblock Repaint: Inpainting using denoising diffusion probabilistic models.
\newblock In \emph{Proceedings of the IEEE/CVF Conference on Computer Vision
  and Pattern Recognition (CVPR)}, 2022.

\bibitem[Ma et~al.(2024)Ma, Lin, Zhang, Ye, Gao, and Liu]{stableflow2024}
Haonan Ma, Weijie Lin, Yuwei Zhang, Yuwei Ye, Ruijia Gao, and Ziwei Liu.
\newblock Stableflow: Progressive flow-guided scene generation from text.
\newblock \emph{arXiv preprint arXiv:2311.16466}, 2024.

\bibitem[Mou et~al.(2024)Mou, Wang, Song, Shan, and Zhang]{mou2024diffedit}
Chong Mou, Xintao Wang, Jiechong Song, Ying Shan, and Jian Zhang.
\newblock {DiffEditor}: Boosting accuracy and flexibility on diffusion-based
  image editing.
\newblock In \emph{Proceedings of the IEEE/CVF Conference on Computer Vision
  and Pattern Recognition}, 2024.

\bibitem[Nichol et~al.(2022)Nichol, Dhariwal, Ramesh, Shyam, Mishra, McGrew,
  Sutskever, and Chen]{nichol2022glide}
Alex Nichol, Prafulla Dhariwal, Aditya Ramesh, Pranav Shyam, Pamela Mishra, Bob
  McGrew, Ilya Sutskever, and Mark Chen.
\newblock {GLIDE}: Towards photorealistic image generation and editing with
  text-guided diffusion models.
\newblock In \emph{Proceedings of the International Conference on Machine
  Learning (ICML)}, pages 16784--16804, 2022.

\bibitem[Radford et~al.(2021)Radford, Kim, Hallacy, Ramesh, Goh, Agarwal,
  Sastry, Askell, Mishkin, Clark, Krueger, and Sutskever]{radford2021learning}
Alec Radford, Jong~Wook Kim, Christopher Hallacy, Aditya Ramesh, Gabriel Goh,
  Sandhini Agarwal, Girish Sastry, Amanda Askell, Pamela Mishkin, Jack Clark,
  Gretchen Krueger, and Ilya Sutskever.
\newblock Learning transferable visual models from natural language supervision
  ({CLIP}).
\newblock In \emph{Proceedings of the International Conference on Machine
  Learning (ICML)}, 2021.

\bibitem[Rombach et~al.(2022)Rombach, Blattmann, Lorenz, Esser, and
  Ommer]{rombach2022high}
Robin Rombach, Andreas Blattmann, Dominik Lorenz, Patrick Esser, and Bj{\"o}rn
  Ommer.
\newblock High-resolution image synthesis with latent diffusion models.
\newblock In \emph{Proceedings of the IEEE/CVF Conference on Computer Vision
  and Pattern Recognition (CVPR)}, pages 10684--10695, 2022.

\bibitem[Singh et~al.(2019)Singh, Natarajan, Shah, Jiang, Chen, Batra, Parikh,
  and Rohrbach]{singh2019textvqa}
Amanpreet Singh, Vivek Natarajan, Meet Shah, Yu Jiang, Xinlei Chen, Dhruv
  Batra, Devi Parikh, and Marcus Rohrbach.
\newblock Towards {VQA} models that can read.
\newblock In \emph{Proceedings of the IEEE/CVF Conference on Computer Vision
  and Pattern Recognition (CVPR)}, pages 8317--8326, 2019.

\bibitem[Song et~al.(2021)Song, Sohl-Dickstein, Kingma, Kumar, Ermon, and
  Poole]{song2021scorebased}
Yang Song, Jascha Sohl-Dickstein, Diederik~P. Kingma, Abhishek Kumar, Stefano
  Ermon, and Ben Poole.
\newblock Score-based generative modeling through stochastic differential
  equations.
\newblock In \emph{International Conference on Learning Representations
  (ICLR)}, 2021.

\bibitem[Tuo et~al.(2024)Tuo, Xiang, He, Gao, and Xie]{tuo2023anytext}
Yuxiang Tuo, Wangmeng Xiang, Jun-Yan He, Yifeng Gao, and Enze Xie.
\newblock {AnyText}: Multilingual visual text generation and editing.
\newblock In \emph{Proceedings of the International Conference on Learning
  Representations (ICLR)}, 2024.

\bibitem[Veit et~al.(2016)Veit, Matera, Neumann, Matas, and
  Belongie]{veit2016cocotext}
Andreas Veit, Tomas Matera, Luk{\'a}{\v{s}} Neumann, Jiri Matas, and Serge
  Belongie.
\newblock {COCO-Text}: Dataset and benchmark for text detection and recognition
  in natural images.
\newblock In \emph{arXiv preprint arXiv:1601.07140}, 2016.

\bibitem[Zhang et~al.(2023)Zhang, Rao, and Agrawala]{zhang2023controlnet}
Lvmin Zhang, Anyi Rao, and Maneesh Agrawala.
\newblock Adding conditional control to text-to-image diffusion models.
\newblock In \emph{Proceedings of the IEEE/CVF International Conference on
  Computer Vision (ICCV)}, 2023.

\bibitem[Zhao and Lian(2024)]{zhao2024udifftext}
Yiming Zhao and Zhouhui Lian.
\newblock {UDiffText}: A unified framework for high-quality text synthesis in
  arbitrary images via character-aware diffusion models.
\newblock In \emph{Proceedings of the IEEE/CVF Conference on Computer Vision
  and Pattern Recognition}, 2024.

\end{thebibliography}
}


\end{document}